\documentclass[letterpaper]{article} 
\usepackage{aaai24}  
\usepackage{times}  
\usepackage{helvet}  
\usepackage{courier}  
\usepackage[hyphens]{url}  
\usepackage{graphicx} 
\usepackage{natbib}  
\usepackage{caption} 
\usepackage{algorithm}
\usepackage{algorithmic}
\newcommand{\mx}{\mathbf{x}}

\newcommand{\ie}{\textit{i.e.}}
\usepackage{amsmath}
\usepackage{amsthm}
\usepackage{booktabs}
\theoremstyle{definition}
\usepackage{amsmath,amsfonts,amsthm} 
\usepackage{amssymb}
\usepackage{color}
\usepackage{subfigure}
\usepackage{multirow}
\usepackage{marvosym}
\usepackage{ifsym}
\usepackage{colortbl}  
\usepackage{xcolor}
\usepackage{array}   
\definecolor{table_row}{rgb}{0.90,0.90,0.91}

\usepackage{newfloat}
\usepackage{listings}
\DeclareCaptionStyle{ruled}{labelfont=normalfont,labelsep=colon,strut=off} 
\floatstyle{ruled}
\newfloat{listing}{tb}{lst}{}
\floatname{listing}{Listing}
\title{Revealing the Proximate Long-Tail Distribution in Compositional Zero-Shot Learning }
\author {
     Chenyi Jiang\textsuperscript{},
    Haofeng Zhang\textsuperscript{ \thanks{Corresponding author}}
}
\affiliations {
     School of Computer Science and Engineering, Nanjing University of Science and Technology, China\\
  \{jiangchenyi, zhanghf\}@njust.edu.cn,
}

\usepackage{bibentry}

\begin{document}

\maketitle

\begin{abstract}
Compositional Zero-Shot Learning (CZSL) aims to transfer knowledge from seen state-object pairs to novel unseen pairs. In this process, visual bias caused by the diverse interrelationship of state-object combinations blurs their visual features, hindering the learning of distinguishable class prototypes. Prevailing methods concentrate on disentangling states and objects directly from visual features, disregarding potential enhancements that could arise from a data viewpoint. Experimentally, we unveil the results caused by the above problem closely approximate the long-tailed distribution. As a solution, we transform CZSL into a proximate class imbalance problem. We mathematically deduce the role of class prior within the long-tailed distribution in CZSL. Building upon this insight, we incorporate visual bias caused by compositions into the classifier's training and inference by estimating it as a proximate class prior. This enhancement encourages the classifier to acquire more discernible class prototypes for each composition, thereby achieving more balanced predictions. Experimental results demonstrate that our approach elevates the model's performance to the state-of-the-art level, without introducing additional parameters. Our code is available at \url{https://github.com/LanchJL/ProLT-CZSL}.
\end{abstract}
\section{Introduction}
\label{sec:intro}
Objects in the world often exhibit diverse states of existence; an \textit{apple} can be \textit{sliced} or \textit{unripe}, while a \textit{building} can be \textit{ancient} or \textit{huge}. Humans have the ability to recognize the composition of the unseen based on their knowledge of seen elements. Even if people have never seen a \textit{green apple} before, they can infer the characteristics of a \textit{green apple} from a \textit{red apple} and a \textit{green lemon}. To empower the machine with this capability, previous work \cite{misra2017red,purushwalkam2019task} propose Compositional Zero-Shot Learning (CZSL), a task aims to identify unseen compositions from seen state-object compositions.

\begin{figure*}[t]
		\centering
		\subfigure 
		{
			\includegraphics[width=0.300\textwidth]{ 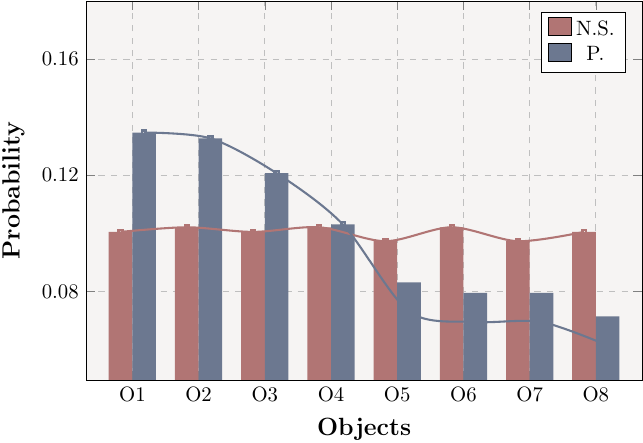}
		}
        \subfigure 
		{
			\includegraphics[width=0.300\textwidth]{ 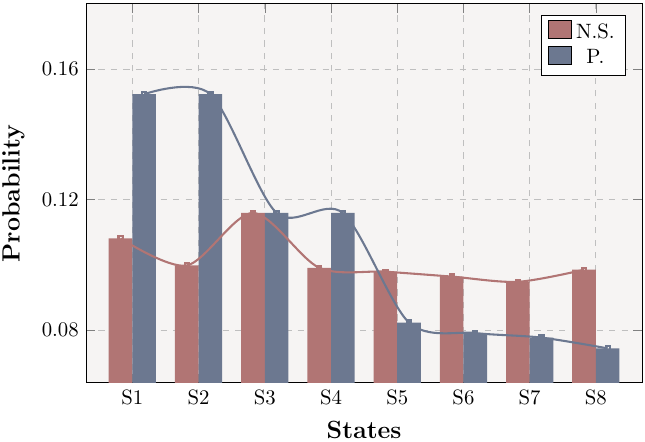}
		}
         \subfigure 
		{
			\includegraphics[width=0.300\textwidth]{ 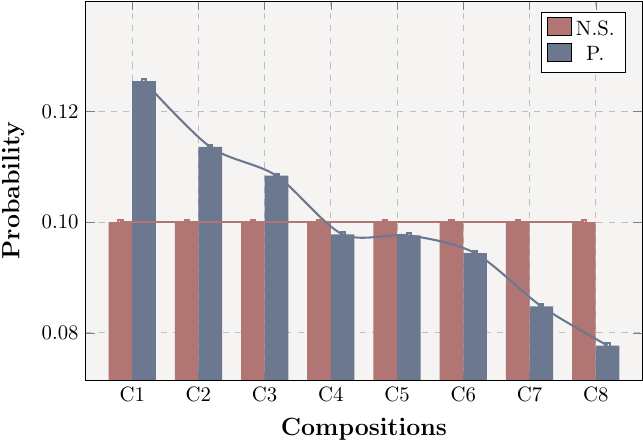}
		}
		\caption{An example of prior and posterior probabilities (predicted by model) of same classes in MIT-States \cite{mit}. N.S. represents the prior probability calculated from the number of samples in each class. P. denotes the average value of posterior probabilities indicating the likelihood of the sample belonging to its class, which is predicted by a MLP with ResNet-18 \cite{resnet} as backbone (same settings as $\mathcal{C}_{o},\mathcal{C}_{s}$ and $\mathcal{C}_{y}$ in Implementation Details and trained via vanilla cross-entropy loss). The classes are selected on the basis of the closest sample size, and shows the results for the object (left), state (middle), and composition (right) classes. All data is simply normalized for ease of presentation.}
		\label{fig1}
	\end{figure*}

However, the combination of state-objects creates a visual bias for a attribute (state or object) in it, hindering the learning of distinguishable class prototypes. In the face of above challenge, early approaches in the domain of CZSL can be categorized into two distinct methods. The first method utilized two independent classifiers to categorize states and objects \cite{misra2017red, li2020symmetry, purushwalkam2019task, li2022siamese}. The second method involved training a common embedding space where semantic and visual features could be projected to reduce the distance between them \cite{naeem2021learning, mancini2021open, mancini2022learning}. Commonly, these studies concentrate on improving the structure of classifiers and investigating alternative architectures. However, minimal research has been conducted considering the problem in terms of data distribution.


We analyze the prior and posterior probabilities associated with attributes (states or objects) and compositions to determine a more suitable solution. Fig. \ref{fig1} illustrates that the class prior follows a distinct trend differing from the posterior probabilities. For instance, even though the model is trained on a comparable number of samples, it demonstrates a low probability of predicting the object labeled as \textit{O5}. This issue also extends to making inferences about compositions, which reminds us of the long-tail distribution or class imbalance \cite{menon2020long,tang2020unbiased,kim2020detecting}. 

We consider that certain samples are infected by the intricate interplay between objects and states within compositions \cite{atzmon2020causal}, leading to significant bias from the ideal class prototype. Consequently, these samples with large visual bias make it difficult for the classifier to fit their intrinsic patterns, results in the inability to form effective classification boundaries. In contrast to class imbalance, we refer to this phenomenon as \textit{`attribute imbalance'} below. The recent methods for CZSL \cite{saini2022disentangling,wang2023learning} synchronize the prediction of visual features to states and objects with the prediction of compositions in the common embedding space, which works as a model ensemble approach. While this design addresses the capability to categorize some classes, the non-interaction among the independent classifiers may lead to incomplete mutual compensation due to potential information gaps.

The identified shortcomings prompted a redesign of the model using the model ensemble approach. Building on the success of logit adjustment in addressing long-tail learning \cite{menon2020long}, this study treats attribute imbalance information as special prior knowledge (In the following we denote by \textit{`attribute prior'}) that approximates the class prior. This attribute prior is derived from the estimation of available samples by two independent classifiers for states and objects. In other words, we construct this prior by modelling the visual bias of states and objects from samples. During the training phase, we incorporate it through logit adjustment into the common embedding space. This approach enables the production of balanced posterior probabilities regarding the poorly-classified classes in Fig. \ref{fig1}, thereby preventing each independent classifier from ineffectively reinforcing the ability to classify the well-classified classes.

Specifically, we reconstructed the CZSL problem from the perspective of mutual information and adjusted the posterior values predicted by the model from the perspective of maximizing mutual information. In addition, we generalize the above attribute prior to the unseen class in order to optimize the lower bound of seen-unseen balanced accuracy \cite{xian2017zero} obtained by \citet{chen2022zero}. We refer to this method as the logit adjustment for \textbf{Pro}ximate \textbf{L}ong-\textbf{T}ail Distribution (ProLT) in CZSL. Unlike previous methods, ProLT does not necessitate introducing additional parameters, yet it significantly enhances the overall CZSL model performance. Our contributions are summarized as follows:

\begin{itemize}
\item In our study, we conduct an analysis of the data distribution in CZSL. We translate the visual bias in compositions into an attribute imbalance and thereby generalize CZSL to a proximate long-tail learning problem. 
\item Our analysis involves a mathematical examination of both the training and inference phases of the model. This enables us to adapt the model's posterior probability based on the attribute prior.
\item Our model enhances the prediction of relationships in compositions without the need for introducing additional parameters. Experimental results on three benchmark datasets demonstrate the effectiveness of our approach.
\end{itemize}

\section{Related Work} \label{sec.related}

\paragraph{Compositional Zero-Shot Learning (CZSL):} Zero-Shot Learning (ZSL) transfers knowledge from seen classes to unseen ones by leveraging attributes \cite{akata2013label, lampert2013attribute, parikh2011relative, frome2013devise, akata2015evaluation}. CZSL \cite{atzmon2020causal, yang2022decomposable,nagarajan2018attributes,wang2019task} builds upon this foundation by incorporating the notion of composition learning \cite{hoffman1984parts}, with its extension primarily relying on the shared semantics of state and object within the composition of both seen and unseen classes.

Initial CZSL methodologies directly classify states and objects, effectively converting the task into a conventional supervised assignment \cite{misra2017red, chen2014inferring, yang2020learning,lu2016visual,li2022siamese}. However, the fusion of state-object pairs led to visual bias in both elements, impeding the acquisition of discernible class prototypes. Numerous subsequent strategies utilize visual-semantic alignment within a common embedding space \cite{naeem2021learning,mancini2022learning,mancini2021open} to grasp the entwined nature of objects and states within compositions. However, this technique is susceptible to domain shift challenges. Recent methodologies typically amalgamate these two models, creating a framework of model ensembles. For instance, \citet{saini2022disentangling} enhances the model's adaptability to unseen classes by disentangling visual features and subsequently reconstituting them for novel classes. Meanwhile, \citet{wang2023learning} introduces conditional state generation to address visual alterations arising from object-state combination. ProLT aligns closely with this paradigm, although with a greater emphasis on direct inquiries into visual bias attributes.

\paragraph{Long-Tailed Classification:} Numerous studies address the issue of imbalanced class distributions, with one prominent approach being posterior modification methods \cite{long1, long3, hou2021detecting, menon2020long, focal}. Within ZSL, \citet{chen2022zero} regards it as an imbalanced challenge involving seen and unseen classes, and then applies regulatory techniques based on logit adjustment. However, this approach does not readily extend to the issue of attribute imbalance in our context. \citet{jiang2022mutual} considers the presence of visual bias in samples re-weighting within the optimization process, but its localization-based weighting strategy ignores the differences between classes. In this study, we introduce advanced logit adjustment strategies theoretically, aiming to enhance the equilibrium of predictions between various classes.

\section{Methodology}\label{Sec.methodology}

\subsection{Task Definition} \label{subsec.taskdefine}
Considering the two disjoint sets $\mathcal{Y}^S$ and $\mathcal{Y}^U$, \ie, $\mathcal{Y}^S \cap \mathcal{Y}^U = \O$. CZSL aims to classify sample $\mx \in \mathcal{X}$ into a composition $y=(s,o) \in \mathcal{Y}$, where $\mathcal{Y} = \mathcal{Y}^S \cup \mathcal{Y}^U$, and samples from $\mathcal{Y}^U$ are unseen during training. $y$ is composed by state $s\in \mathcal{S}$ and object $o \in \mathcal{O}$, $\mathcal{S}$ and $\mathcal{O}$ are sets of states and objects. Samples from $\mathcal{Y}^S$ and $\mathcal{Y}^{U}$ share the same objects $o$ and states $s$, but their compositions $(s,o)$ are different. Define the visual space $\mathcal{X}\subseteq \mathbb{R}^{d_x}$ and $d_x$ is the dimension of the space, $\mathcal{X}$ can be divided into $\mathcal{X}^S$ and $\mathcal{X}^U$ based on whether their samples belong to seen classes. We can define the train set as $\mathcal{D}_{seen}=\{(\mx,y) | \mx \in \mathcal{X}^S, y \in \mathcal{Y}^S \}$ and an unseen set for evaluation of methods which is $\mathcal{D}_{unseen}=\{(\mx,y)|\mx\ \in \mathcal{X}^U,y\in \mathcal{Y}^U\}$. We employ the Generalized ZSL setup defined in \citet{xian2017zero}, which requires both seen and unseen classes involves in testing.

\subsection{Empirical Analysis on Model Ensemble} \label{subsec.empirical_analysis}
	\begin{table}[H]
 \Huge
		\centering
        	\resizebox{0.47\textwidth}{!}{
		\begin{tabular}{ccccccc}
			\toprule
			Method                & M.E.         &Sta.   & Obj.    & S.& U.  & HM  \\
			\midrule
			\multirow{2}{*}{MLP \cite{mancini2021open}}   & F     & 27.9 & 31.8 & 25.3 & 24.6&16.4 \\
			&   T &  27.9&  32.0&  29.8&  24.5&  17.9 \\
    \cmidrule(lr){1-7}
			\multirow{2}{*}{GCN \cite{naeem2021learning}}  & F   &27.9 & 32.0&28.7& 25.3&17.2\\
	&   T  &  28.3&  33.4&  28.9&  26.0&  18.8\\
    \cmidrule(lr){1-7}
   I.C. &-&25.3&24.8&19.3&19.0&12.0\\
			\bottomrule
		\end{tabular}
        }
		\caption{The results of methods that incorporate a composition classifier, along with the addition of two classifiers for states and objects on top of them (\ie, model ensemble), are presented. I.C. indicates only two independent classifiers are used. M.E. indicates the utilization of model ensemble in the methods, where F denotes false and T denotes true. The metrics in the table are defined in Evaluation Protocol.}
\label{tab.empirical}
	\end{table}

For the problem of approximate long-tailed distributions caused by visual bias in CZSL, ensemble-based methods have demonstrated exceptional performance in CZSL \cite{saini2022disentangling, wang2023learning}. Typically, this approach combines the predictions of two models to produce the final prediction. The first model consists of two independent classifiers $C_o$ and $C_s$ for objects and states. The second model is a composition classifier $C_y$. The process of the model can be viewed as inputting the samples into three classifiers to estimate the posterior probabilities:
	\begin{equation}
	\begin{split}
    \begin{aligned}
	&p(s|\mx) = \operatorname{softmax}[C_s(\mx)], p(o|\mx) = \operatorname{softmax}[C_o(\mx)],\\ &p(y|\mx) = \operatorname{softmax}[C_y(\mx)],\\ &\hat{p}(y|\mx) =  \delta p(y|\mx)+ (1-\delta)\left [p(s|\mx)+p(o|\mx)\right ],
    \end{aligned}
	\end{split}
    \label{eq1}
	\end{equation}
where $p(s|\mx),p(o|\mx)$ and $p(y|\mx)$ are posterior probability from classifiers, $\hat{p}(y|\mx)$ is the final posterior probabilities. $\delta$ is a weight factor. $\mathcal{C}_{y}(\mx)$ denotes the logits for class $y$ based on sample $\mx$, and $\mathcal{C}_{s}(\mx)$, $\mathcal{C}_{o}(\mx)$ are similarly defined.

As demonstrated in Tab. \ref{tab.empirical}, augmenting two additional posterior estimates $p(o|\mx)$ and $p(s|\mx)$ to $p(y|\mx)$ can significantly enhance CZSL results. However, only relying solely on $p(o|\mx)$ and $p(s|\mx)$ does not enable accurate estimation, this suggests the improvement in results is not due to the introduction of superior classifiers. Consequently, we can deduce the subsequent conjectures: The effectiveness of ensemble-based methods emanates from incorporating $\mathcal{C}_{s}$ and $\mathcal{C}_{o}$, aiding in the classification of compositions that encounter a relative disadvantage within $\mathcal{C}_y$. While attribute imbalances vary across states, objects, and compositions, all three elements might not simultaneously experience large visual bias for a particular class. Based on these preliminary studies, we can posit that effective classification of classes with large visual bias within common embedding spaces requires information compensation. In our study, we directly estimate visual bias as compensation described above. Considering that the visual bias generated by state-object combination is difficult to eliminate directly, we try to introduce it as an attribute prior into the training process from the classifier. In the following, we detail this process.

\subsection{From the Perspective of Mutual Information} \label{subsec_mutal}
Let us first consider the problem from a simple CZSL approach based on common embedding spaces like \cite{mancini2021open,naeem2021learning}. The optimization objective of these methods can be viewed as the maximum likelihood:
	\begin{equation}
		\operatorname{argmin}_{\theta}\mathbb{E}_{(\mx,y)\sim \mathcal{D}_{seen}}[-logp(y|\mx)],
        \label{eq_objective}
	\end{equation}
where $p(y|\mx)$ is defined in Eq. \ref{eq1}, which denotes the distribution of compositions predicted by the model. $\theta$ denotes the model parameters.

Given the characteristics of CZSL, where each sample is associated with two labels, $s$ and $o$, there is a conditionality between the two in the setup of the dataset, \ie,
	\begin{equation}
		p(y)=p(s,o)=p(o)p(s|o),
        \label{eq_condition}
	\end{equation}
$p(y)$ and $p(o)$ denotes class prior of class $y$ and object $o$, and $p(s|o)$ is conditional class prior of $s$ and $o$. 

Inspired by \citet{kexuefm-7615},  we look at the above issues through the perspective of mutual information \cite{kraskov2004estimating}, we have:
 	\begin{equation}
 	\begin{split}
		&I(Y;X)
  \approx \mathbb{E}_yD_{KL}[p(y |\mx)\| p(y)]\\
  =&\sum_{\mx,y}p(\mx,y)log\frac{p(y|\mx)}{p(y)}\\
    =&\sum_{\mx,y}p(\mx,y)log\frac{p(y|\mx)}{p(o)p(s|o)},
        \label{eq_mutual_information}
        \end{split}
	\end{equation}
where $X$ and $Y$ are discrete random variables corresponding to $\textbf{x}$ and $y$, respectively, and $S$ and $O$ are similarly defined. $D_{KL}$ represents the Kullback-Leibler divergence, while $p(\mx,y)$ represents the joint probability of the class $y$ and the visual feature $\textbf{x}$. Due to the real posterior probability between $y$ and $\mx$ is unknown, we use $p(y|\mx)$ as an approximation. We can interpret the optimization of maximum likelihood as follows, based on the posterior term in  Eq. \ref{eq_mutual_information},
	\begin{equation}
    log\frac{p(y|\mx)}{p(o)p(s|o)}\sim \mathcal{C}_{y}(\mx),
        \label{eq_f_log}
	\end{equation}
which can be transfer to:
	\begin{equation}
 	\begin{split}
    logp(y|\mx) \sim \mathcal{C}_{y}(\mx)+log{p(o)p(s|o)},
        \label{eq_f_log_transfer}
        \end{split}
	\end{equation}
here, $\mathcal{C}_{y}(\mx)$ represents the logits for class $y$, defined in Eq. \ref{eq1}, $\sim$ denotes approximately equal. The expression on the right-hand side is re-normalized using the $\operatorname{softmax}$ function, \ie,
 	\begin{equation}
  	 \resizebox{0.97\linewidth}{!}{$
 	\begin{split}
  &-logp(y|\mx) \\
  &\sim log \left [1+\sum_{o_i\neq o}\sum_{s_j \neq s}\frac{p(o_i)p(s_j|o_i)}{p(o)p(s|o)}e^{\mathcal{C}_{\hat{y}}(\mx)- \mathcal{C}_{y}(\mx)}\right ]\\
  &\sim  log \left [1+\sum_{o_i\neq o}\sum_{s_j \neq s}\left (\frac{p(o_i)p(s_j|o_i)}{p(o)p(s|o)} \right )^{\eta}e^{ \mathcal{C}_{\hat{y}}(\mx)- \mathcal{C}_{y}(\mx)}\right ],
        \label{eq_after_softmax}
        \end{split}
        $}
	\end{equation}
where $\hat{y}=(s_i,o_i)$, and $\eta$ is an adjustment factor. Eq. \ref{eq_after_softmax} demonstrates that by incorporating the class prior $p(s|o)$ and $p(o)$ for state $s$ and object $o$, we can optimize the model's mutual information. Consequently, we approach the CZSL problem from the perspective of mutual information.

\subsection{Estimating the Attribute Prior}\label{subsec.from_posteri_view}
The above idea comes from the logits adjustment \cite{menon2020long} introduced to address class imbalance \cite{johnson2019survey,japkowicz2002class}, which demonstrate that the inclusion of a class prior enhances the maximization of mutual information, and we generalize it to CZSL task.

As stated in Introduction, we undertake the transformation of CZSL into an approximate long-tailed distribution issue caused by visual bias from state-object combinations. Our argument centers on the proposition that attribute imbalance within CZSL contributes to an approximate form of class imbalance, since visual bias hinders reduces the distinguishability of some of the samples. Therefore, exclusive reliance on the class prior is inadequate. Building upon this rationale, we propose to use the attribute prior to assume the function of the class prior within the long-tailed distribution, serving as an approximation.

We propose incorporating the model's conditional posterior probabilities as an approximation for this scenario. We continue to denote it as the `prior' due to its function as a prior probability during the training process, despite being computed using posterior probability. Since attribute imbalance cannot be directly quantified from the dataset, we simulate it by utilizing the posterior probability of the additional classifiers, for $\mx$ and its corresponding $s,o$, we have:
	\begin{equation}
 	\begin{split}
    \hat{p}(s) = \mathbb{E}_{\mx\sim p(\mx)}[p(s|\mx)],
    \hat{p}(o) = \mathbb{E}_{\mx\sim p(\mx)}[p(o|\mx)],
        \label{eq_estimate}
        \end{split}
	\end{equation}
where $\mx \in \mathcal{D}_{seen}$, $p(s|\mx)$ and $p(o|\mx)$ are defined in Eq. \ref{eq1}, which are posterior probabilities from $\mathcal{C}_{s}$ and $\mathcal{C}_{o}$, we use their predicted expectations for all training samples as a special attribute prior. From this we can replace the class prior in Eq. \ref{eq_after_softmax} with following item:
	\begin{equation}
k(s,o)= \operatorname{softmax}\left[ \sigma(s,o)\hat{p}(s)\hat{p}(o)\right ],
        \label{eq_final_sigma_introduce}
	\end{equation}
where $\sigma(s,o)$ is a function used to model the conditional nature of the composition, \ie,
 \begin{equation}\label{eq_onehot}
\sigma(s,o)=\left\{
\begin{array}{rcl}
&1& (s,o) \in \mathcal{Y}^{S}\cup \mathcal{Y}^{U},\\
&0 &else.\\
\end{array} \right. 
\end{equation}
From this we obtain the final objective function according to Eq. \ref{eq_after_softmax}: 
 	\begin{equation}
\mathcal{L}_{cls}= log \left [1+\sum_{o_i\neq o}\sum_{s_j \neq s}\left (\frac{k(s_j,o_i)}{k(s,o)} \right )^{\eta}e^{ \mathcal{C}_{\hat{y}}(\mx)- \mathcal{C}_{y}(\mx)}\right ].
        \label{eq_final_after_softmax}
	\end{equation}

 \subsection{Logit Adjustment for Inference} \label{subsec.inference}
Due to the introduction of unseen classes in the inference phase we need to make additional adjustments. CZSL usually measures model performance in terms of $\mathcal{A}^{H}$ which denotes Harmonic Mean (HM) accuracy:
	\begin{equation}
	\mathcal{A}^H=2/(\frac{1}{\mathcal{A}^S}+\frac{1}{\mathcal{A}^U}),
    \label{eq_ah}
	\end{equation}
 where $\mathcal{A}^{S}$, $\mathcal{A}^U$ denote seen and unseen accuracy.  \citet{chen2022zero} provides a lower bound of HM, below we briefly describe its conclusions. For HM's lower bound we have:
	\begin{equation}
	\mathcal{A}^{H}\ge 1/\mathbb{E}_{\mathbf{x}\sim p(\mathbf{x})}\frac{|\mathcal{Y}|p(\mathcal{Y})p(y|y\in \mathcal{Y})}{q(\mathcal{C}_{out}=y|\mathbf{x})p(y|\mathbf{x})},
    \label{eq_lowerbound}
	\end{equation}
where $q(\mathcal{C}_{out}=y|\mx)$ represents the probability of predicting class $y$ using our model. The set $\mathcal{Y}$ can be either $\mathcal{Y}^{S}$ or $\mathcal{Y}^{U}$, $p(y|y\in \mathcal{Y})$ represents the conditional class prior, and $|\mathcal{Y}|p(\mathcal{Y})$ can be seen as a hyper-parameter that quantifies the differences between seen and unseen classes. Considering that the gap between the domains of seen and unseen classes in CZSL is not significant, we can simply treat $|\mathcal{Y}|p(\mathcal{Y})$ as an ignorable constant in the following process.

Finding the Bayesian optimum for $\mathcal{A}^{H}$ is difficult. However, it is possible to maximize its lower bound, which is equal to minimizing the upper bound of its inverse, \ie, the denominator term of Eq. \ref{eq_lowerbound} is minimized if:
\begin{equation}
		\tilde{y}=\operatorname{argmax}_{y}\left [\mathcal{C}_{y}(\mx)+\eta logp(y|y\in \mathcal{Y})\right ],
        \label{eq_infer_1}
	\end{equation}
where $\eta$ is from Eq. \ref{eq_final_after_softmax}, $\tilde{y}$ is the predicted label for sample $\mx$. For conditional class prior $p(y|y\in \mathcal{Y}^{S})$, which represents the true class frequency when $y$ belong to seen classes. Following Eq. \ref{eq_final_after_softmax}, we similarly replace the prior with the attribute prior estimate in Eq. \ref{eq_final_sigma_introduce} here, which is:
\begin{equation}
p(y | (s,o) \in \mathcal{Y}^{S}) := k(s,o), \quad (s,o) \in \mathcal{Y}^{S},
\label{eq_seen}
\end{equation}
and the attribute prior of unseen classes are not available to the model, we model it here using a combination of the estimation from Importance Sampling \cite{importancesampling} with the attribute prior from seen samples, which can be denoted as:
\begin{equation}
p(y|y\in \mathcal{Y}^{U}):=k(s,o)+\frac{\hat{k}_{\mx}(s,o)}{\lambda k(s,o)}, \quad (s,o) \in \mathcal{Y}^{U},
        \label{eq_importance}
	\end{equation}
where $\frac{1}{\lambda}$ is a hyper-parameter denotes the distribution of $\mx$. The above results are re-transformed into probability distributions in the actual calculation. And $\hat{k}_{\mx}(s,o)$ is instance-based conditional posterior probability:
\begin{equation}
\hat{k}_{\mx}(s,o)=\operatorname{softmax} \left [ \sigma(s,o)p(s|\mx)p(o|\mx)\right ],
        \label{eq_khat}
	\end{equation}
the aforementioned setup arises because during testing, we are unable to provide posterior probabilities $\hat{k}_{\mx}(s,o)$ from multiple samples simultaneously. Furthermore, Importance Sampling results in significant variance when the number of samples is insufficient. To address this, we attempt to augment it by leveraging seen attribute prior.
 
ProLT makes inferences during testing phase based on Eq. \ref{eq_infer_1}, our aim is to integrate local information during testing with the prior derived from seen classes, to address the disparities between seen and unseen classes. With Eq. \ref{eq_infer_1} ProLT theoretically achieves the best overall accuracy.

	\begin{figure}[t]
		\centering
		\subfigure
		{
			\includegraphics[width=0.42\textwidth]{ 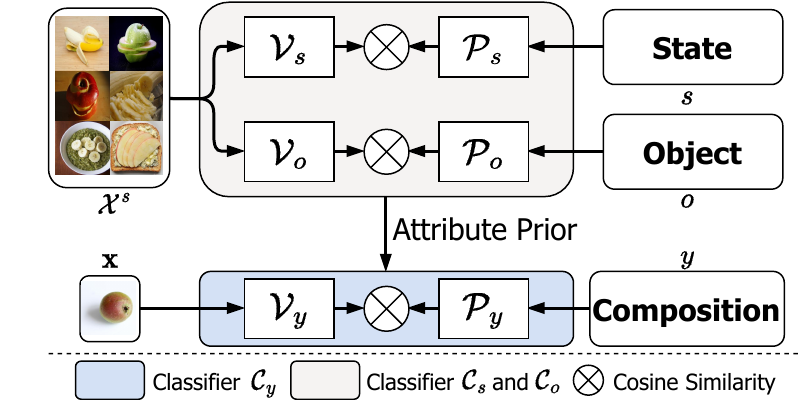}
		}
		\caption{A brief demonstration of ProLT's training stage (detailed in Method Overview). $\mathcal{X}^s$ is the set of seen visual features, we obtain the attribute prior according to Eq. \ref{eq_estimate}.}
        
		\label{fig_method}
	\end{figure}

\subsection{Method Overview}\label{subsec.overview}
This section provides a concise summary of the aforementioned methods. Our approach, illustrated in Fig. \ref{fig_method}, involves training two independent classifiers denoted as $\mathcal{C}_{s}$ and $\mathcal{C}_{o}$. These classifiers are implemented using prototype learners, namely $\mathcal{P}_{s}$ and $\mathcal{P}_{o}$, and visual embedders $\mathcal{V}_{o}$, $\mathcal{V}_{s}$, to determine the prototypes of states and objects, \ie,
\begin{equation}
\mathcal{C}_{s}(\mx):=\frac{cos(\mathcal{V}_{s}(\mx),\mathcal{P}_{s}(s))}{\tau},\mathcal{C}_{o}(\mx):=\frac{cos(\mathcal{V}_{o}(\mx),\mathcal{P}_{o}(o))}{\tau},
        \label{eq_prototype}
	\end{equation}
where $\tau$ is the temperature. These classifiers are trained with vanilla cross-entropy loss:
\begin{equation}
\mathcal{L}_{ic}= log[1+\sum_{s'\neq s}e^{\mathcal{C}_{s'}(\mx)-\mathcal{C}_{s}(\mx)} ][1+\sum_{o'\neq o}e^{\mathcal{C}_{o'}(\mx)-\mathcal{C}_{o}(\mx)} ].
        \label{eq_ce}
	\end{equation}
Once the classifiers reach a specific training stage, we calculate the attribute prior using Eq. \ref{eq_estimate}, and employ the loss function $\mathcal{L}_{cls}$ from Eq. \ref{eq_final_after_softmax} for training the classifier $\mathcal{C}_{y}$ for compositions:
\begin{equation}
\mathcal{C}_{y}(\mx):=\frac{cos(\mathcal{V}_{y}(\mx),\mathcal{P}_{y}(y))}{\tau},
        \label{eq_prototype_common}
	\end{equation}
where $\mathcal{P}_{y}$ is the prototype learner for compositions and $\mathcal{V}_{y}$ is a visual embedder. After training, the model uses Eq. \ref{eq_infer_1} for inference.
\section{Experiments} \label{sec.experiments}
\subsection{Datas\label{subsec.datasets}}
There are numerous recent approaches to compositionality research, and three datasets have been primarily employed for evaluation: MIT-States \cite{mit}, UT-Zappos \cite{utzappos}, and C-GQA \cite{naeem2021learning}. We utilized a standardized evaluation dataset for a reasonable comparison with previous methods.

\textbf{MIT-States} presents a considerable challenge, consists of 53,753 images. It comprises 115 state classes, 245 object classes, and 1,962 compositions. In the total compositions, there are 1,262 seen compositions, and 700 compositions remain unseen. \textbf{UT-Zappos} is a collection of 50,025 images that focuses on various forms of footwear. It consists of 12 object classes and 16 state classes which is a fine-grained dataset, yielding 116 compositions, of which 83 are seen. \textbf{C-GQA} is introduced by \citet{naeem2021learning}, which encompasses a wide variety of real-world common objects. It comprises 413 states, 674 objects, and over 27,000 images, along with more than 9,000 compositions, consisting of 5,592 seen and 1,932 unseen compositions.

\subsection{Evaluation Protocol} \label{subsec.metrics}
The setting of GZSL \cite{xian2017zero} requires both seen and unseen compositions during testing. We report the \textbf{best accuracy} of \textbf{seen classes (best seen)}, the \textbf{unseen class (best unseen)}, and its \textbf{harmonic accuracy (HM)}. In order to measure the performance on attribute learning, we report the \textbf{best accuracy} of \textbf{states} \textbf{(best sta)} and \textbf{objects (best obj)}. Building upon the research of \cite{naeem2021learning} and \cite{wang2019task}, we calculate the \textbf{Area Under the Curve (AUC)} by comparing the accuracy on seen and unseen compositions with various bias terms. 

\begin{table*}[ht] 
		\centering
   \huge
  		\resizebox{\textwidth}{!}{
			\begin{tabular}{@{}c|l|cccccc|cccccc|cccccc}
				\toprule
    		&\multirow{2}{*}{Methods}&\multicolumn{6}{c|}{MIT-States} & \multicolumn{6}{c|}{UT-Zappos}& \multicolumn{6}{c}{C-GQA} \\ 
     
       && \multicolumn{1}{c}{AUC}&HM&S.&U.&Sta.&Obj.&\multicolumn{1}{c}{AUC}&HM&S.&U.&Sta.&Obj.&\multicolumn{1}{c}{AUC}&HM&S.&U.&Sta.&Obj.\\
				\midrule
     \multicolumn{1}{c|}{\multirow{9}{*}{$\dagger$}} &LE+ \shortcite{misra2017red}&2.0&10.7&15.0&20.1&23.5&26.3&25.7&41.0&53.0&61.9&41.2&69.2&0.8&6.1&18.1&5.6&-&-\\
       & AttOp \shortcite{nagarajan2018attributes}&1.6&9.9&14.3&17.4&21.1&23.6&25.9&40.8&59.8&54.2&38.9&69.6&0.7&5.9&17.0&5.6&-&-\\
&TMN \shortcite{purushwalkam2019task}&2.9&13.0&20.2&20.1&23.3&26.5&29.3&45.0&58.7&60.0&40.8&69.9&1.1&7.5&23.1&6.5&-&-\\
&SymNet \shortcite{li2020symmetry}&3.0&16.1&24.4&25.2&26.3&28.3&23.9&39.2&53.3&57.9&40.5&71.2&2.1&11.0&26.8&10.3&-&-\\
&CompCos \shortcite{mancini2021open}&4.5&16.4&25.3&24.6&27.9&31.8&28.7&43.1&59.8&62.5&44.7&73.5&2.6&12.4&28.1&11.2&-&-\\
&CGE \shortcite{naeem2021learning}&5.1&17.2&28.7&25.3&27.9&32.0&26.4&41.2&56.8&63.6&45.0&73.9&2.3&11.4&28.1&10.1&-&-\\
&SCEN \shortcite{li2022siamese}&5.3&18.4&29.9&25.2&28.2&32.2&32.0&47.8&63.5&63.1&47.3&75.6&2.9&12.4&28.9&12.1&13.6&27.9\\
&Co-CGE \shortcite{mancini2022learning}
&5.1&17.5&27.8&25.2&-&-&29.1&44.1&58.2&63.3&-&-&2.8&12.7&29.3&11.9\\
&OADis \shortcite{saini2022disentangling}&5.9&18.9&31.1&25.6&28.4&33.2&30.0&44.4&59.5&65.5&46.5&75.5&-&-&-&-&-&-\\
&DECA \shortcite{yang2022decomposable}&5.3&18.2&29.8&25.5&-&-&31.6&46.3&62.7&63.1&-&-&-&-&-&-&-&-\\
&CANet \shortcite{wang2023learning}&5.4&17.9&29.0&26.2&30.2&32.6&33.1&47.3&61.0&66.3&48.4&72.6&\textbf{3.3}&\textbf{14.5}&30.0&13.2&17.5&22.3\\
& \textbf{ProLT} (\textbf{Ours})& \textbf{6.0}& \textbf{19.3}& 30.9& 26.5& 29.5& 34.2& \textbf{33.4}& \textbf{49.3}& 62.7& 64.0& 46.1& 74.2& 3.2& 14.4& 32.1& 13.7& 17.8& 32.5\\
\midrule
\multicolumn{1}{c|}{\multirow{3}{*}{$\ddagger$}}&CSP \shortcite{csp}&19.4&36.3&46.6&49.9&-&-&33.0&46.6&66.2&64.2&-&-&6.2&20.5&26.8&28.8&-&- \\

&DFSP \shortcite{dfsp}&20.8&37.7&52.8&47.1&-&-&36.0&47.2&66.7&71.7&-&-&10.5&27.1&38.2&32.0 \\

& \textbf{ProLT} (\textbf{Ours})& \textbf{21.1}& \textbf{38.2}& 49.1& 51.0& 49.8& 59.0& \textbf{36.1}& \textbf{49.4}& 66.0& 70.1& 52.6& 79.4& \textbf{11.0}& \textbf{27.7}& 39.5& 32.9& 24.9& 50.1\\
\bottomrule
			\end{tabular}
   }
		\caption{The SoTA comparisons on three datasets. We compare ProLT with others on AUC, best HM, best sta (Sta.), best obj (Obj.), best seen (S.) and best unseen (U.). $\dagger{}$ denotes ResNet-based methods and $\ddagger$ denotes CLIP-based methods. The best AUC and HM for ResNet-based methods and CLIP-based methods are shown in bold.}
	\label{tab_results}
\end{table*}
\subsection{Implementation Details} \label{subsec.implementation}
Below we present the details of the implementation of ProLT on ResNet-18 \cite{resnet}. 

\paragraph{Visual Representations and Semantic:} In line with prior methods, we employed ResNet-18 pre-trained on ImageNet \cite{2009ImageNet} to extract 512-dimensional visual features from the images. For semantic information, we utilized GloVe \cite{glove} to extract attribute names as 300-dimensional word vectors.

\paragraph{Implementations and Hyper-Parameters:} For three prototype learner $\mathcal{P}_{s},\mathcal{P}_{o}$ and $\mathcal{P}_{y}$ are GloVe connects with  two Fully Connected (FC) layers with ReLU \cite{relu} following the first layer. And the three visual embedders $\mathcal{V}_{s},\mathcal{V}_{o}$, and $\mathcal{V}_{y}$ are also two FC layers with ReLU  and Dropout \cite{dropout}. All FCs embed the input features in 512 dimensions and the hidden layer is 1024 dimensions. The overall model is trained using the Adam optimizer \cite{kingma2014adam} on NVIDIA GTX 2080Ti GPU, and it is implemented with PyTorch \cite{paszke2019pytorch}. We set the learning rate as $5\times10^{-4}$ and the batchsize as 128. We train the $\mathcal{C}_{s},\mathcal{C}_o$ and $\mathcal{C}_{y}$ with an early-stopping strategy, it needs about 400 epochs on MIT-States, 300 epochs on UT-Zappos and 400 epochs on C-GQA. For hyper-parameters, we set $\tau$ as $0.1,0.1,0.01$, $\eta$ as $1.0,1.0,1.0$ and $\lambda$ as $50,10,100$ for MIT-States, UT-Zappos, and C-GQA, respectively.

\subsection{Compared with State-of-the-Arts} \label{subsec.compare}
ProLT is mainly compared with recent methods using fixed ResNet-18 as backbone with the same settings. We also compared ProLT with the CLIP-based approaches \cite{dfsp,csp} after using CLIP \cite{clip} to learn visual and semantic embeddings. The comparison results are shown in Tab. \ref{tab_results}.

The results demonstrate that ProLT achieves a new state-of-the-art performance when using ResNet-18 as backbone on the MIT-States, UT-Zappos, and C-GQA. Specifically, our method achieves the highest AUC of $6.0\%$ on MIT-States, surpassing CANet by $0.6\%$. On the UT-Zappos, we achieve the highest HM of $49.3\%$, outperforming CANet by $2.0\%$. Although ProLT has a slight disadvantage on the C-GQA dataset, it remains competitive with the state-of-the-art methods, achieving an HM of $14.4\%$. As for the CLIP-based approaches, ProLT has produced remarkable outcomes. Unlike DFSP, our method avoids the incorporation of extra self-attention or cross-attention mechanisms. Despite this, we excel across all three datasets, attaining an HM of $38.2\%$ on MIT-States and $49.4\%$ on UT-Zappos. These results underscore the compatibility of ProLT when combined with CLIP.

	\begin{table}[t]
		\centering
		\resizebox{0.470\textwidth}{!}{
			\begin{tabular}{lccccccc}
				\toprule
				\multirow{1}{*}{Method} & \multirow{1}{*}{Prior}   &AUC&HM&S.&U.&Sta.&Obj. \\ 
				\midrule    
				\multirow{3}{*}{GCN}           & N. &28.4&45.0&58.9&60.0&43.4&70.0\\
				& C.P.     & 29.6   & 46.2   & 58.9   & 61.3   & 43.9  & 71.8\\
    & A.P.     &32.3&48.4&61.7&62.4&45.9&73.2 \\
				\midrule    
				\multirow{3}{*}{FC}     &N. &32.3&46.8&61.1&64.8&44.1&72.3 \\
				&C.P.     &32.0&47.9&62.2&62.1&44.2&73.7 \\ 
    &A.P.     &33.4&49.3&62.7&64.0&46.1&74.2 \\ 
				\bottomrule
			\end{tabular}
		}
		\caption{A comparison of different priors for Eq. \ref{eq_final_after_softmax} and Eq. \ref{eq_infer_1} on UT-Zappos. GCN: GCN is used as the prototype learner, FC: FC layers are used as the prototype learner. N. represents no prior is introduced, using pure ensemble method, C.P. represents class prior from datasets is utilized, and A.P. denotes attribute prior is incorporated.}
		\label{tab.ablation1}
	\end{table}
 	\begin{table}[t]
		\centering
  \huge
		\resizebox{0.470\textwidth}{!}{
			\begin{tabular}{lcccccccc}
				\toprule
				\multirow{1}{*}{Method} & \multirow{1}{*}{$\eta=0$}&$p=0$   &AUC&HM&S.&U.&Sta.&Obj. \\ 
				\midrule    
				\multirow{3}{*}{GCN.}           & $\checkmark$ &$\times$&29.4&44.2&59.9&63.0&43.4&70.0\\
    & $\times$ &$\checkmark$& 31.3  & 47.3   & 60.8  & 60.5   & 43.7 & 73.6 \\
       & $\times$ &$\times$&32.3&48.4&61.7&62.4&45.9&73.2 \\
				\midrule    
				\multirow{3}{*}{FC.}      & $\checkmark$ &$\times$ &28.9&44.1&60.1&62.9&44.1&73.2\\
    & $\times$ &$\checkmark$& 32.7   &48.3   &61.8  & 64.6   &45.8   & 74.1\\
    & $\times$ &$\times$ &33.4&49.3&62.7&64.0&46.1&74.2 \\ 
				\bottomrule
			\end{tabular}
		}
		\caption{Ablation results for each component on UT-Zappos. $p=0$ deontes we remove the prior in Eq. \ref{eq_infer_1}, $\checkmark$ indicates setting $p$ or $\eta$ to $0$, and $\times$ indicates the opposite.}
		\label{tab.ablation2}
	\end{table}

 	\begin{figure*}[t]
		\centering
		\subfigure
		{
			\includegraphics[width=0.93\textwidth]{ 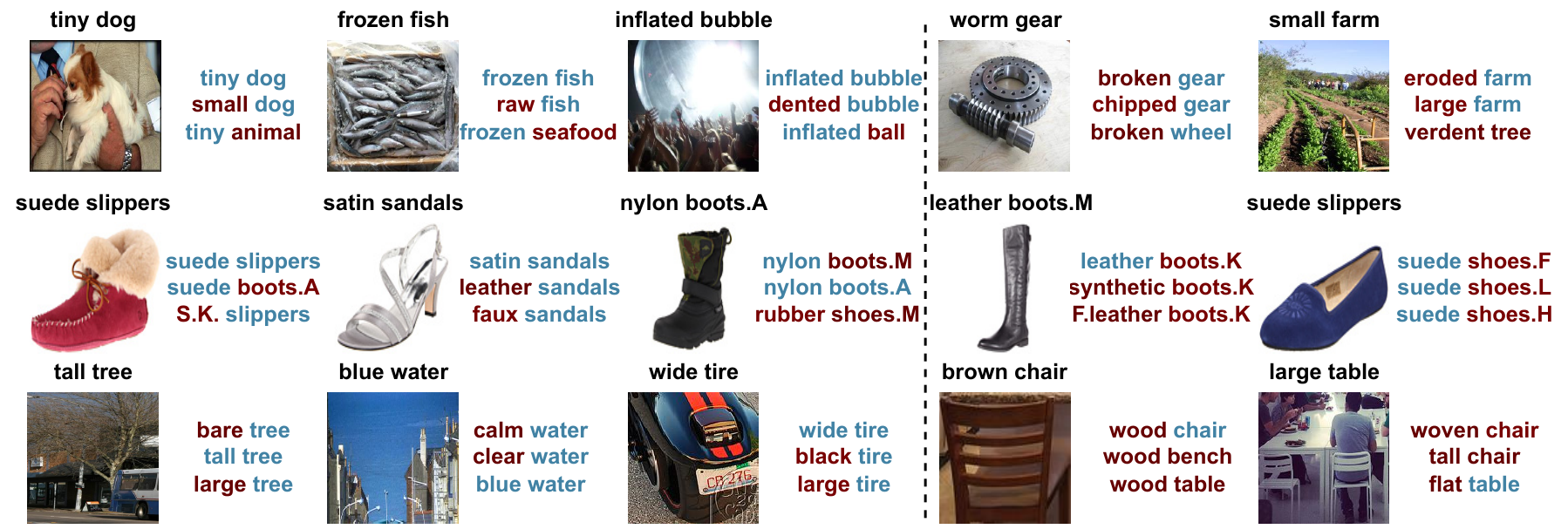}
		}
		\caption{Qualitative results on MIT-States (first row), UT-Zappos (second row) and C-GQA (third row), where the left part contains the top-3 results contains correct predicts, and the rights contains the top-3 predicts do not contain correct predicts. The label is indicated in black above the image, with correctly predicted results indicated in blue and incorrect ones in red.}
		\label{fig_qua}
	\end{figure*}
\subsection{Ablation Study} \label{subsec.ablation}
In this section, we verify that each of these modules plays an active role by ablating each of its parts on UT-Zappos with ResNet-18. The results are shown in Tab. \ref{tab.ablation1} and Tab. \ref{tab.ablation2}.
\paragraph{Attribute Prior versus Class Prior:} As mentioned above, we use the attribute prior in place of the class prior due to the attribute imbalance. To further validate this, we replaced Eq. \ref{eq_final_after_softmax} and Eq. \ref{eq_infer_1} using class prior, shown in Tab. \ref{tab.ablation1}. To make the results more robust, we tested two different prototype learners, \ie, the GCN from the CGE \cite{naeem2021learning} and the FC layers. The results in Tab. \ref{tab.ablation1} indicate that incorporating a class prior yields improvements over the baseline. We attribute this enhancement mainly to \citet{menon2020long}, the class sizes of datasets are not solely identical. However, ProLT exhibits a substantial advantage over the other methods, which demonstrates the more dominant influence of potential attribute imbalances in CZSL.

\paragraph{Effect of Components:} 
We eliminate the effects of each component by adjusting the hyper-parameters $\eta$ in Eq. \ref{eq_final_after_softmax} and the attribute prior in Eq. \ref{eq_importance} to verify the role played by each component. In Tab. \ref{tab.ablation2}, we set $\eta$ to 0 to convert Eq. \ref{eq_final_after_softmax} to a vanilla cross entropy loss and the inference phase is converted to same as CGE. For $p=0$, we remove the attribute prior in inference phase. We also tested on both prototype learners. We can observe that each part of the ablation leads to a decrease in outcome, with $\eta=0$ being the most significant. This reflects the effectiveness of our method.

  \begin{figure}[t]
	\centering
		\subfigure
		{
			\includegraphics[width=0.23\textwidth]{ 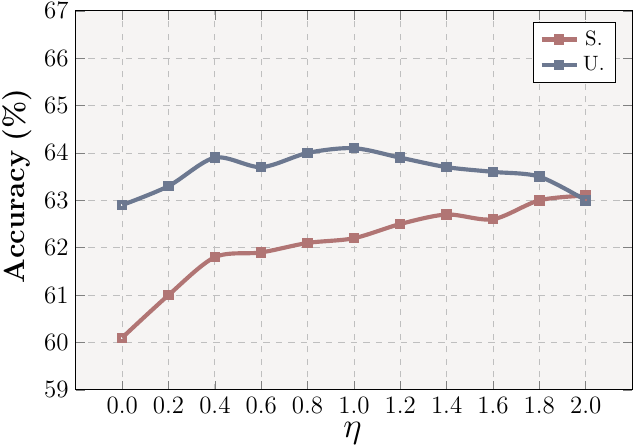}
		}
  		\subfigure
		{
			\includegraphics[width=0.21\textwidth]{ 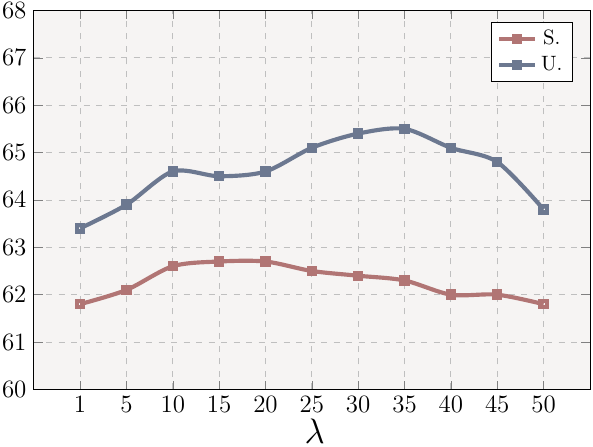}
		}
  		\subfigure
		{
			\includegraphics[width=0.23\textwidth]{ 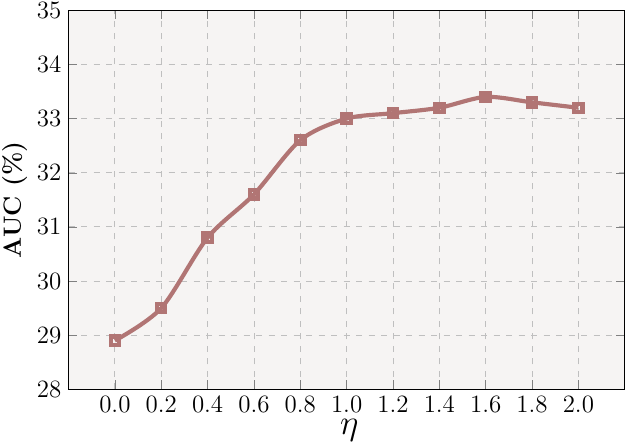}
		}
  		\subfigure
		{
			\includegraphics[width=0.21\textwidth]{ 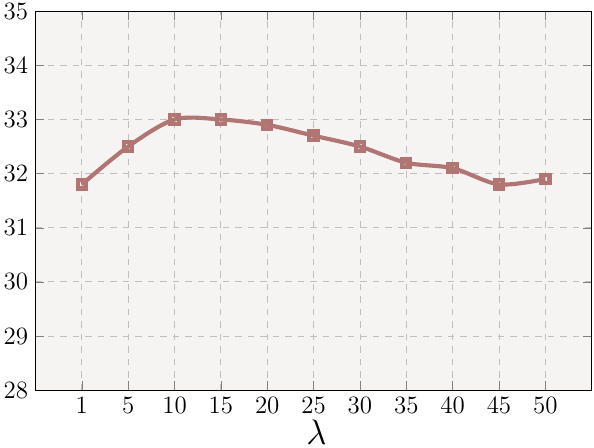}
		}
	\caption{Influence of hyper-parameters on UT-Zappos about best seen (S.), best unseen (U.) and AUC.}
	\label{fig_hyper}
\end{figure}
\subsection{Hyper-Parameter Analysis}\label{subsec.hyper_parameters}
Our method primarily comprises the subsequent hyper-parameters: 1) logit-adjusting factor ($\eta$), and 2) factor about the sample distribution ($\lambda$). We test on the UT-Zappos under various hyper-parameters based on ResNet-18, shown in Fig.~\ref{fig_hyper}. For $\eta$, the best AUC are observed when $\eta=1.6$, and the gap between seen and unseen begins to decrease as $\eta$ increases. Concerning $\lambda$, the outcomes are documented within the interval $\lambda \in [1.0, 50.0]$ with increments about 5.0. The pinnacle value for the seen class is observed at $20.0$, and $35.0$ for unseen class. Overall, these hyper-parameter settings yield results characterized by minimal fluctuations, thus underscoring the robustness of our methodology.

\subsection{Qualitative Results} \label{subsec.qualitative}
Qualitative results for unseen compositions, accompanied by the top-3 predictions when we use ResNet-18 as backbone, are displayed in Fig. \ref{fig_qua}. Concerning MIT-States, we argue that certain erroneous predictions as partially justifiable. For instance, the phrase \textit{tiny dog}, for which the model's incorrect predictions involve \textit{small dog} and \textit{tiny animal}, exhibits a high degree of semantic similarity. A similar phenomenon can be observed for the \textit{brown chair} in C-GQA. For UT-Zappos, ProLT's limitation in fine-grained classification persists. An illustrative example is the outcomes for \textit{leather boot.M}, our approach encounters challenges in making nuanced differentiations within the category of boots.

\section{Conclusion}
This paper presents from an experimental analysis aimed at revealing the concealed proximate long-tail distribution issue within CZSL. In our work, CZSL is transformed into an underlying proximate class imbalance problem, and the logit adjustment technique is employed to refine the posterior probability for individual classes. Diverging from conventional methods for handling long-tailed distributions, the introduced attribute prior is derived from the model's sample estimation of visual bias. Experimental results demonstrate that our approach attains state-of-the-art outcomes without necessitating the introduction of supplementary parameters.

\section*{Acknowledgements}
This work was supported by National Natural Science Foundation of
China (NSFC) under the Grant No. 62371235.

\bibliography{aaai24}

\clearpage
\noindent \section{Appendix}
\renewcommand\thefigure{A.\arabic{figure}} 
\renewcommand\theequation{A.\arabic{equation}} 
\renewcommand\thetable{A.\arabic{table}} 

\section{Supplementary Experiments and Details with Vision-Language Model}
\subsection{Implementation Details (Supplemental)}
In this section, we provide details of the setup of ProLT when using CLIP to learn visual and semantic embeddings.

\paragraph{Visual Representations and Semantic:} We employ the pretrained CLIP Vit-L/14 model as both our image and test encoder. Regarding semantics, we employ a learnable soft prompt \textit{[v1][v2][v3][state][object]}, following the approach of \citet{dfsp}, where \textit{[v1][v2][v3]} represent the learnable content. To embed attributes such as state or object, we compute the average embedding value for each composition containing the corresponding state or object.

\paragraph{Implementations and Hyper-Parameters: }The three prototype learners, $\mathcal{P}_{s},\mathcal{P}_{o}$, and $\mathcal{P}_{y}$, adhere to the configuration detailed in Sec. 4.3, except for the omission of GloVe \cite{glove}. Similarly, the three visual embedders, $\mathcal{V}_{s},\mathcal{V}_{o}$, and $\mathcal{V}_{y}$, remain consistent with the specifications in Sec. 4.3. We train the entire model using the Adam optimizer \cite{kingma2014adam} on two NVIDIA GTX 3090 GPUs, while configuring the batch size as 16. The other hyper-parameter configurations remain consistent with those in main text.

\subsection{Ablation Study with CLIP}
Following Sec. 4.5, we conducted an identical experiment on CLIP to validate the effectiveness of ProLT. As demonstrated in Tab. \ref{subtab.ablation}, we compare the outcomes on UT-Zappos \cite{utzappos} under three scenarios: without incorporating any priors but using a model-ensemble method, with the inclusion of class priors, and with the inclusion of attribute priors. Similarly, the results demonstrate the beneficial impact of incorporating the attribute prior. In comparison to the direct utilization of the class prior, our approach leads to a rise of $1.9\%$ in AUC and $2.0\%$ in HM.

Moreover, we conduct a comparative analysis by removing the attribute prior during both the training and testing phases. Referencing Tab. \ref{subtab.ablation2}, when $\eta=0$, indicating our methods is changed to a simple common embedding space method like \citet{naeem2021learning}, which led to a significant drop in results. A significant enhancement is observed when these are combined, similar to the findings in Tab. 4. Collectively, the aforementioned experiments substantiate the favorable impact of ProLT on CLIP.

\section{Additional Experiments and Further Information}
In this section we add some detailed information from Sec. 4 as well as perform some additional experiments. All experiments are performed with ResNet-18 \cite{resnet} as the backbone.
\subsection{Training Details}
\paragraph{Early Stopping:} 
As mentioned in Sec. 3.6, ProLT requires that $\mathcal{C}_{s}$ and $\mathcal{C}_{o}$ be trained together first using $\mathcal{L}_{ic}$. In this process we simply employ an early stopping strategy on the validation set. We trained these module for a maximum of 50 epochs and use AUC for early-stopping. After $\mathcal{C}_{s}$ and $\mathcal{C}_{o}$ training is complete, it starts outputting attribute priors and co-training with $\mathcal{C}_y$. This process we adopt the same early stopping strategy on the validation set. We set the maximum of 1000 epochs and also use AUC for early-stopping.

\begin{table}[t]
		\centering
		\resizebox{0.482\textwidth}{!}{
			\begin{tabular}{lccccccc}
				\toprule
				\multirow{1}{*}{Method} & \multirow{1}{*}{Prior}   &AUC&HM&S.&U.&Sta.&Obj. \\ 
				\midrule    
				\multirow{3}{*}{CLIP}           & N. &33.6&46.5&63.4&69.2&50.9&79.5\\
				& C.P.     & 34.2   & 47.4   & 64.1   & 68.1   &51.2  &77.8\\
    & A.P.     &36.1&49.4&66.0&70.1&52.6&79.4\\				
				\bottomrule
			\end{tabular}
		}
        	\vspace{-0.5ex}
		\caption{A comparison of different priors for Eq. 11 and Eq. 14 when using CLIP as image and text encoder on UT-Zappos. \textbf{N.:} no prior is introduced, using pure ensemble method. \textbf{C.P.:} class prior from datasets is utilized. \textbf{A.P.:} attribute prior is incorporated.}
        	\vspace{-2ex}
		\label{subtab.ablation}
	\end{table}

  	\begin{table}[t]
		\centering
		\resizebox{0.482\textwidth}{!}{
			\begin{tabular}{lcccccccc}
				\toprule
				\multirow{1}{*}{Method} & \multirow{1}{*}{$\eta=0$}&$p=0$   &AUC&HM&S.&U.&Sta.&Obj. \\ 
				\midrule    
				\multirow{3}{*}{CLIP}           & $\checkmark$ &$\times$& 31.7   &45.9  & 62.7 &66.2  &48.8& 75.7\\
    & $\times$ &$\checkmark$& 35.1  & 47.8   & 65.0  & 69.3   & 52.1 &79.9 \\
       & $\times$ &$\times$ &36.1&49.4&66.0&70.1&52.6&79.4\\
				\bottomrule
			\end{tabular}
		}
        	\vspace{-2ex}
		\caption{Ablation results for each component on UT-Zappos when using CLIP as image and text encoder. $p=0$ deontes we remove the prior in Eq. 14, $\checkmark$ indicates setting $p$ or $\eta$ to $0$, and $\times$ indicates the opposite.}
        	\vspace{-2.0ex}
		\label{subtab.ablation2}
	\end{table}

\begin{table}[ht]
\begin{center}
\scalebox{0.90}{
	\centering
    \begin{tabular}{@{}l|cccccc}
    \toprule
          \textit{Word Embeddings} &\multicolumn{1}{c}{\textit{AUC}}&HM&S.&U.&Sta.&Obj.\\
				\midrule
GloVe&33.4&49.3&62.7&64.0&46.1&74.2  \\
 Word2Vec&32.4&48.8&63.1&64.9&45.6&75.0\\
 Fasttext&32.5&48.4&63.0&62.6&45.4&74.9\\
     GloVe+Word2Vec&33.0&49.4&63.1&63.8&45.4&74.1\\
     Fasttext+Word2Vec&33.3&49.0&63.2&64.9&45.5&75.5\\
	    	\bottomrule
    \end{tabular}	
}
	\caption{Results on UT-Zappos using different word embedding. }
	\label{subtab_wordembedding}
 \end{center}
 \vspace{-3ex}
 \end{table}

 \begin{table}[ht]
\begin{center}
\scalebox{0.90}{
	\centering
    \begin{tabular}{@{}c|cccccc}
    \toprule
          \textit{Dimension} &\multicolumn{1}{c}{\textit{AUC}}&HM&S.&U.&Sta.&Obj.\\
				\midrule
256&32.7&47.9&61.4&63.5&45.0&74.0 \\
 512&33.1&48.9&62.2&63.7&45.0&73.6\\
 1024&33.4&49.3&62.7&64.0&46.1&74.2\\
     2048&32.3&48.0&61.7&63.7&44.9&74.3\\
     4096&32.2&48.3&61.9&64.4&46.3&74.7\\
	    	\bottomrule
    \end{tabular}	
}
	\caption{Reults on UT-Zappos using different hidden layer settings.}
	\label{subtab_dimension}
 \end{center}
 \vspace{-3ex}
 \end{table}

  \begin{figure}[t]
	\centering
		\subfigure
		{
			\includegraphics[width=0.22\textwidth]{ 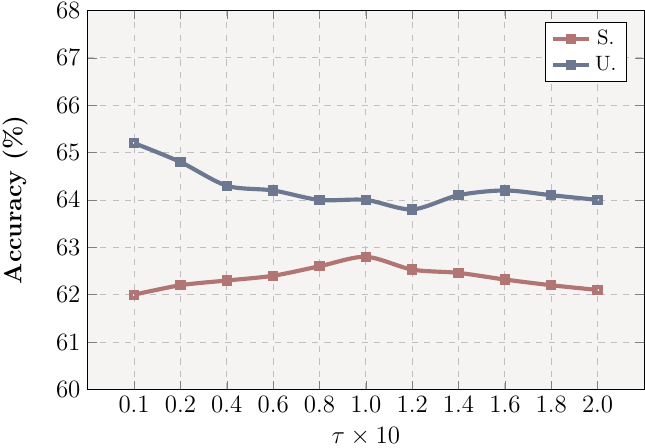}
		}
  		\subfigure
		{
			\includegraphics[width=0.22\textwidth]{ 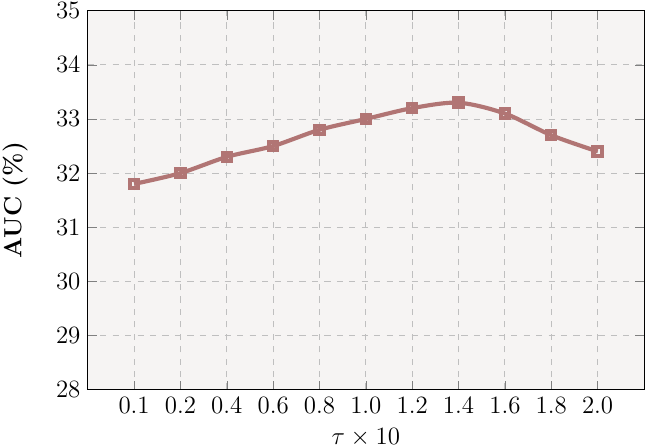}
		}
	\caption{Influence of $\tau$ on UT-Zappos about best seen (S.), best unseen (U.) and AUC.}
	\label{subfig_hyper}
 \vspace{-2.5ex}
\end{figure}
 
\paragraph{Hyper-Parameter Selection:} 
Hyper-parameter selection involves grid-search on the validation set. For architectural parameters, we explore the 1) hidden layer count for $\mathcal{V}_{s}$, $\mathcal{V}_{o}$, and $\mathcal{V}_{y}$ within the range ${0,1,2}$, and 2) hidden layer count for $\mathcal{P}_{s}$, $\mathcal{P}_{o}$, and $\mathcal{P}_{y}$ within the same range. Concerning optimization, such as learning rate, we adopt the configuration from \citet{naeem2021learning} without extensive modifications. For the remaining hyper-parameters, we search for $\eta\in [0.0,2.0]$ with a step of 0.2, $\lambda \in [5,50]$ with a step of 5 for UT-Zappos, and $\lambda \in [10,200]$ with a step of 10 for MIT-States and C-GQA. Additionally, we perform a search for $\tau \in [0.02,0.2]$ with an increment of 0.02, encompassing the value $\tau=\{0.005,0.01\}$. For the choice of word embedding, we search the word embedding of 1) GloVe \cite{glove}, 2) Word2Vec \cite{mikolov2013distributed}, 3) Fasttext \cite{fasttext}, 4) GloVe+Word2Vec and (5) Fasttext+Word2Vec.

 	\begin{table}[t]
		\centering
		\resizebox{0.482\textwidth}{!}{
			\begin{tabular}{lcccccccc}
				\toprule
				\multirow{1}{*}{Method} & \multirow{1}{*}{\textit{sp} $=0$}&$\lambda\rightarrow \infty$.   &AUC&HM&S.&U.&Sta.&Obj. \\ 
				\midrule    
				\multirow{3}{*}{GCN.}           & $\checkmark$ &$\checkmark$& 31.3  & 47.3   & 60.8  & 60.5   & 43.7 & 73.6 \\
				& $\checkmark$ &$\times$ &31.8&47.9&61.9&62.3&45.9&72.8\\
    & $\times$ &$\checkmark$& 31.7&48.1&61.8&63.0&45.3&73.4\\
       & $\times$ &$\times$&32.3&48.4&61.7&62.4&45.9&73.2 \\
				\midrule    
				\multirow{3}{*}{FC.}      & $\checkmark$ &$\checkmark$ & 32.7   &48.3   &61.8  & 64.6   &45.8   & 74.1\\
				& $\checkmark$ &$\times$ & 32.8   &48.5  &62.0  & 64.0   & 46.0   &74.3 \\
    & $\times$ &$\checkmark$& 32.9  &49.0   &62.5  &63.2   &45.2  & 74.6\\
    & $\times$ &$\times$ &33.4&49.3&62.7&64.0&46.1&74.2 \\ 
				\bottomrule
			\end{tabular}
		}
        	\vspace{-2ex}
		\caption{Ablation results for different inference method. \textit{sp} $=0$ denotes we set the seen attribute prior in Eq. 16 to 0. And $\lambda \rightarrow \infty$ denotes we remove the probability from Importance Sampling. $\checkmark$ indicates true, and $\times$ indicates false. }
		\label{subtab.ablation3}
	\end{table}

 \begin{figure}[t]
	\centering
		\subfigure
		{
			\includegraphics[width=0.46\textwidth]{ 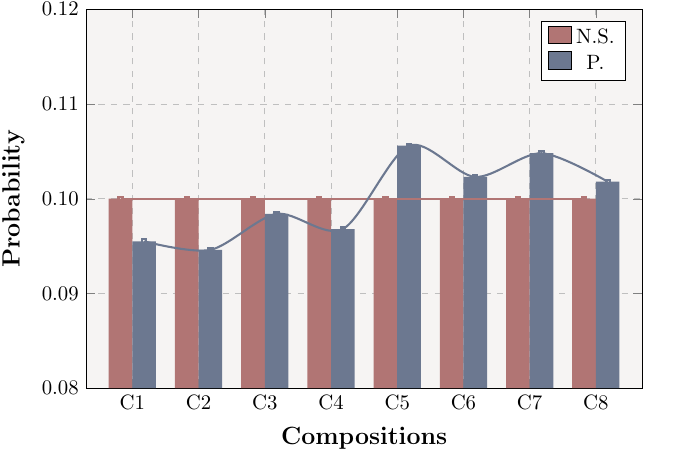}
		}

	\caption{An example of posterior and prior probabilities for various compositions using our method, where \textbf{N.S.} denotes the class prior, and the \textbf{P.} denotes the probabilities of $p(y|\mx)$. Our adjusted posterior provides a more balanced distribution compared to Fig. 1.}
	\label{sub_figre_ana}
\end{figure}

\subsection{Further Experiments of Hyper-Parameters}
This section delves into the analysis of the effects arising from the subsequent hyper-parameter configurations: 1) temperature $\tau$, 2) hidden layer dimensions, and 3) word embeddings. Illustrated in Fig. \ref{subfig_hyper}, we present the outcomes achieved across various $\tau$ values on UT-Zappos. The peak AUC emerges at $\tau=0.14$ and the difference between seen and unseen is minimized at $\tau=0.1$. Likewise, we present outcomes utilizing diverse word embeddings on UT-Zappos, detailed in Tab. \ref{subtab_wordembedding}. ProLT excels when employing GloVe, yet generally, variations in word vectors exhibit minimal impact. Concerning the varying dimension configurations, we document the outcomes obtained using dimensions ${256,512,1024,2048,4096}$ for the hidden layers in three classifiers, as indicated in Tab. \ref{subtab_dimension}. Notably, we discern that a hidden layer dimension of 1024 consistently yields optimal results. However, when employing dimensions of 2048 or 4096, we posit that the inferior performance could result from the propensity of higher-dimensional hidden layers to manifest overfitting on seen classes.

\subsection{Explanation of Importance Sampling}
Eq. 16 employs Importance Sampling to estimate the attribute prior for unseen classes. The specifics of this approach are outlined in this section. During this procedure, we introduce an auxiliary proposal distribution to aid in creating an approximate estimation, \ie, the distribution of the seen attribute prior $k(s,o)$. Therefore, the estimation of the prior for unseen classes can be represented as follows:
	\begin{equation}
\frac{1}{n}\sum_{i}^{n}\frac{p(\mx_i)\hat{k}_{\mx}(s,o)}{k(s,o)}.
        \label{subeq_is}
	\end{equation}
In Eq. 16, $p(\mx_i)$ is replaced by $\lambda$, which is a hyper-parameter. This is owing to the unavailability of direct access to the data distribution for the test set. As the posterior can be obtained only for individual samples during testing, we set $n$ to 1 in practice. This approach yields significant variance due to an inadequate sample size. Consequently, we posit that it should be integrated with the another information.
\subsection{Further Ablation Study on Inference}
In this section, we validate the effectiveness of the approach on UT-Zappos that combining the seen attribute prior with Importance Sampling, as detailed in Sec. 3.5. Tab. \ref{subtab.ablation3} presents the outcomes where we nullify the probability estimated by Importance Sampling via setting $\lambda \rightarrow \infty $ in Eq. 16, and the results when we set the seen attribute prior to $0$. It is worth noting that with seen attribute prior set to 0, we generalize the probability of Importance Sampling to the seen class for consistency. We conducted experiments on two embedding functions to ensure robustness following Tab. 3. From the table, we can observe that the introduction of the two respectively brings about an improvement in AUC, HM relative to the baseline, while it is not significant in the rest of the metrics. In addition, the combination of the two usually creates complementarities, suggesting that they are not mutually exclusive.

\subsection{Why Our Method Works}
As shown in Fig. \ref{sub_figre_ana}, we display the adjusted posterior $p(y|\mx)$ for the same compositions in Fig. 1. It is evident that ProLT yields a more balanced distribution stemming from $\mathcal{C}_{y}$. Furthermore, it becomes apparent that $\mathcal{C}_y$ exhibits a preference for compositions characterized by significant visual bias in Fig. 1. But the incorporation of the prior, as defined in Eq. 14, can mitigates this distinction.

This aspect reflects ProLT: enhancing the model's grasp of fundamental visual-semantic relationships during the training phase through the maximization of mutual information. Utilizing posterior probability adjustment, ProLT achieves classification by harmonizing both a prior and a posterior during the inference phase.

\end{document}